\RequirePackage{amsmath}
\documentclass[runningheads]{llncs} 

\usepackage{graphicx} 
\usepackage{times}
\usepackage{xcolor}
\usepackage{soul}
\usepackage[utf8]{inputenc}
\usepackage{placeins}
\usepackage{subfigure}
\usepackage{amssymb}
\usepackage{amstext}
\usepackage{mathdots}
\usepackage{stmaryrd}
\usepackage{mathtools} 
\usepackage{braket}
\usepackage{url}
\usepackage[ruled]{algorithm2e}
\usepackage{lmodern}
\usepackage[T1]{fontenc}
\usepackage{textcomp}
\usepackage{tikz}
\usepackage{svg}
\usepackage{amsmath}
\usetikzlibrary{matrix}
\usepackage{float}

\usepackage{amsopn}

\begin{document}

\title{
Lifelong Learning Starting From Zero\thanks{Supported by the Torsten S\"oderberg Foundation \"O110/17. Corresponding author: Claes Stranneg\aa rd. Email: \email{claes.strannegard@chalmers.se}}
}

\author{
Claes Stranneg\aa rd\inst{1,2}
\and 
Herman Carlstr\"om \inst{1} \and 
Niklas Engsner \inst{2} \and 
Fredrik M\"akel\"ainen \inst{2}\and 
Filip Slottner Seholm \inst{1}  \and 
Morteza Haghir Chehreghani\inst{1}
}
\authorrunning{C. Stranneg\aa rd  et al.}

\institute{Department of Computer Science and Engineering\\
Chalmers University of Technology, Gothenburg, Sweden
\and
Dynamic Topologies Sweden AB,
Gothenburg, Sweden
}

\maketitle              

\begin{abstract}
We present a deep neural-network model for lifelong learning inspired by several forms of neuroplasticity. The neural network develops continuously in response to signals from the environment. In the beginning the network is a blank slate with no nodes at all. It develops according to four rules: (i) \emph{expansion}, which adds new nodes to memorize new input combinations; (ii) \emph{generalization}, which adds new nodes that generalize from existing ones; (iii) \emph{forgetting}, which removes nodes that are of relatively little use; and (iv) \emph{backpropagation}, which fine-tunes the network parameters. 
We analyze the model from the perspective of accuracy, energy efficiency, and versatility and compare it to other network models, finding better performance in several cases. 
 

\end{abstract}
Animals need to respond rapidly and appropriately to all kinds of changes in their environment. To stay alive, they must make sufficiently good decisions at every moment. With few exceptions, they learn from experience; their decision-making improves over time. That requires effective mechanisms for adding, modifying, removing, and using memories. 


Memories are arguably only useful to the extent they contribute to better decision-making in future: e.g.,
memories of vital resources that can be exploited again;
memories of dangers that need to be avoided; 
memories formed recently; and 
memories used relatively often.

 
The ability to learn continuously by incorporating new knowledge is called \textit{lifelong learning} \cite{PARISI201954}: 
sensory data is available via a continuous data stream; 
that data comes without any division into e.g. training set and test set; 
it comes without division into tasks; and
sensory input at two consecutive time steps tends to be similar. Within computer science, lifelong learning is often contrasted with learning in \textit{batch mode} where the entire data set is available from the start. 


Today, deep-learning models can outperform humans on a number of tasks; see e.g. \cite{dlsurvey}. When it comes to lifelong learning and general intelligence, however, the success of deep learning has been modest at best. 
In contrast, insects like the honeybee and fruit fly excel at lifelong learning and adaptation to new environments \cite{GREENSPAN2004707}. These animals have several mechanisms of \textit{neuroplasticity} for altering their nervous systems in response to changes in the environment \cite{power2017neural}.

The present research was guided by the idea that neural networks with static architectures lack the flexibility needed for effective lifelong learning.
Section \ref{Related} summarizes research in lifelong learning based on neural networks. Section \ref{section:Model} presents our dynamic model LL0. Section \ref{Results} analyzes LL0 from the perspective of accuracy, energy consumption, and versatility. Section \ref{section:Conclusion} draws some conclusions.

\section{Related work}
\label{Related}
Lifelong learning constitutes a long-standing, central problem in machine learning \cite{PARISI201954,Hassabis2017}. Many current neural-network-based learning methods assume that all training data is available from the beginning and do not consider lifelong learning. That said, several models for lifelong learning are based on neural networks. 
\textit{Catastrophic forgetting} is a crucial aspect of lifelong learning \cite{MCCLOSKEY1989109,French1999CatastrophicFI,Soltoggio2018BornTL} that can lead to abrupt deterioration in performance. To get around it, biologically inspired computational methods integrate new knowledge while preventing it from dominating old knowledge \cite{DitzlerRAP15,mermillod2013stability}. The consequent trade-off is referred to as the \textit{stability-plasticity dilemma} \cite{Grossberg1982}.

Various solutions have been proposed. As the network sequentially learns multiple tasks, weight protection \cite{kirkpatrick2017overcoming} counteracts catastrophic forgetting by safeguarding weights that have been important previously. Regularization techniques \cite{Goodfellow-et-al-2016} constrain the update of neural networks to prevent catastrophic forgetting \cite{Li2018LearningWF}. Pruning can be used toward the same end \cite{wolfe2017incredible,8114708}. Both regularization techniques and pruning reduce network size while improving generalization.
The neural-network models developed for these purposes can have fixed or dynamic architectures. With fixed architectures, adaptation to  new knowledge is achieved via parameter updates that penalize parameter updates to avoid catastrophic forgetting. \cite{zenke2017continual} presents an example of such a model with ``synaptic'' intelligence.

Among the earliest dynamic models is the \textit{cascade-correlation} architecture \cite{fahlman1990cascade}, which adds one hidden neuron at a time while freezing the network to avoid catastrophic forgetting. \textit{Progressive} neural networks \cite{rusu2016progressive} add new layers of neurons progressively while blocking changes to those parts of the network trained on earlier data. Other incremental methods exist, based e.g. on incremental training of an auto-encoder; new neurons are added in response to a high rate of failure with the new data \cite{pmlr-v22-zhou12b} or based on reconstruction error \cite{Draelos2017NeurogenesisDL}. AdaNet \cite{cortes2017adanet} gradually extends its network by evaluation and selection among candidate sub-networks. A \textit{dynamically expandable} network \cite{Lee2018LifelongLW} expands via network split/duplication operations, retraining the old network only when necessary. 
Lifelong learning has been applied to such domains as autonomous learning and robotics. Learning agents are continuously exposed to new data from the environment \cite{cdep.12282,Krueger2009FlexibleSH} in a strategy markedly different from classical learning performed on finite, prepared data. 
Lifelong learning is in no way limited to deep neural-network models: consider the methods  used for language \cite{Mitchell2015NeverEndingL} and topic modeling \cite{Chen2014TopicMU}.

\section{The LL0 model}
\label{section:Model}

The supervised-learning model LL0 adds and removes nodes and connections dynamically through four network-modification mechanisms, each inspired by a different form of neuroplasticity: 
(i) \emph{backpropagation} which adjusts parameters, inspired by synaptic plasticity \cite{draganski2008training}; 
(ii) \emph{extension}, which adds new nodes, inspired by neurogenesis \cite{kandel2000principles}; 
(iii) \emph{forgetting}, which removes nodes, inspired by programmed cell death \cite{oppenheim1991cell}; and (iv) \emph{generalization}, which abstracts from existing nodes, inspired by synaptic pruning \cite{paolicelli2011synaptic}. LL0 thus models four forms of neuroplasticity rather than one (i), as in standard deep learning, or two (i+ii), as in the dynamic approaches mentioned above. 

LL0 receives a continuous stream of data points $(x,y)$, where $x$ and $y$ are vectors of real numbers with fixed dimensions. The model maintains a neural network that starts without any nodes or connections and develops continuously. 
Algorithm \ref{algo:main} shows the main loop; the following subsections add details.
\begin{algorithm}
receive the first data point $(x,y)$\\
form $|x|$ input nodes and $|y|$ output nodes\\
\While{true}{
compute network output $\hat{y}$ produced by input $x$\\
\eIf{$prediction(\hat{y})\neq y$}{generalization\\
extension}{backpropagation}
forgetting \\
receive a new data point $(x,y)$}
\caption{Main loop of LL0.}
\label{algo:main}
\end{algorithm}

\noindent
LLO's neural network consists of four node types:
\begin{description}
    \item[input nodes] with the identity function as their activation function;
    \item[output nodes] with softmax as their activation function;
    \item[value nodes] with a Gaussian activation function and two parameters, ($\mu, \sigma$), used for storing values; and
    \item[concept nodes] with a sigmoid activation function and one bias parameter, used for forming the neural counterparts of conjunctions (though training can turn them into something quite different!).
\end{description}
All concept nodes are directly connected to all output nodes. Concept nodes may also have any number of outgoing connections to value nodes. All incoming connections to concept nodes are from value nodes. Each value node has one incoming connection, which originates from either a concept or input node. They have one outgoing connection, always to a concept node.

\subsection{Extension}
\begin{figure}[ht]
  \begin{center}
    \includegraphics[width=0.45\textwidth]{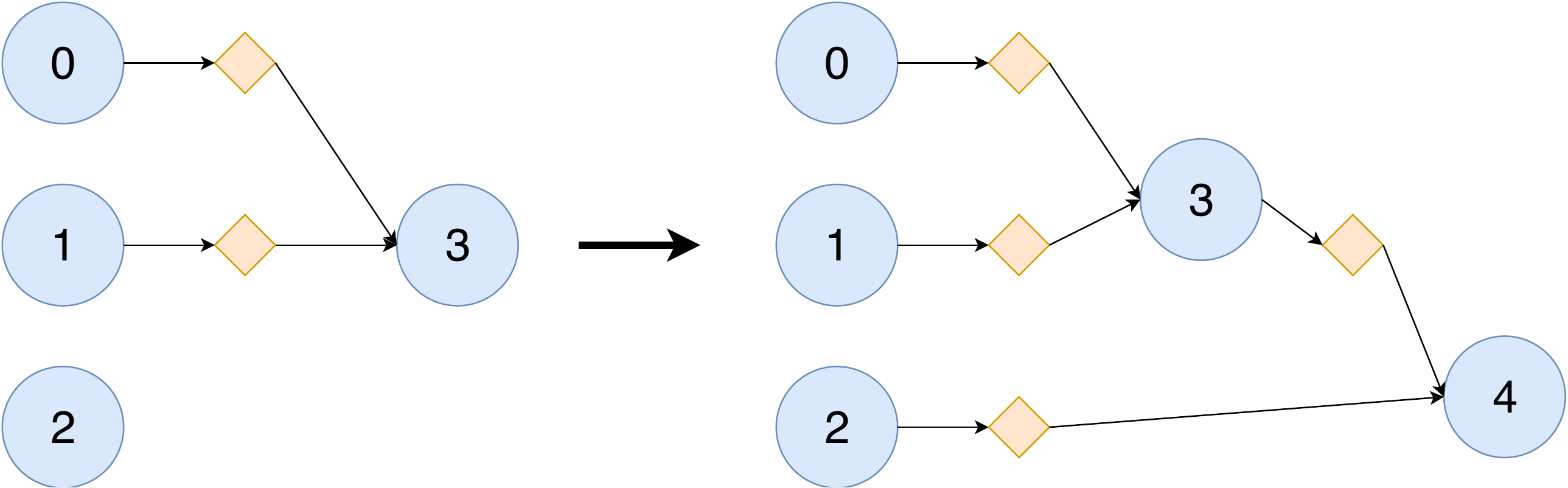}
    \end{center}
    \caption{Illustration of the extension rule. Yellow diamonds represent value nodes, blue circles other nodes. Assuming that nodes 2 and 3 are in the extension set, a new concept node 4 is added along with two new value nodes.}
    \label{fig:1_extend}
\end{figure}

When LL0 makes an incorrect classification, the \textit{extension rule} is triggered. Then an \textit{extension set} is formed. This set consists of all concept nodes and input nodes, whose activation is above a certain threshold and whose position in the network is as deep as possible. Thus no node in the extension set has a downstream node that is also in the extension set. 
The extension rule essentially connects each node of the extension set to a value node and then connects those value nodes to a concept node, as illustrated in Figure \ref{fig:1_extend}. 

\noindent
The parameters are set so that each value node stores the present activation of its parent node and the concept node resembles an AND-gate. 
The concept node is then connected to all output nodes and the weights of those connections are set so that one-shot learning is ensured.

\begin{figure}[ht]
  \centering
    \includegraphics[width=0.3\textwidth]{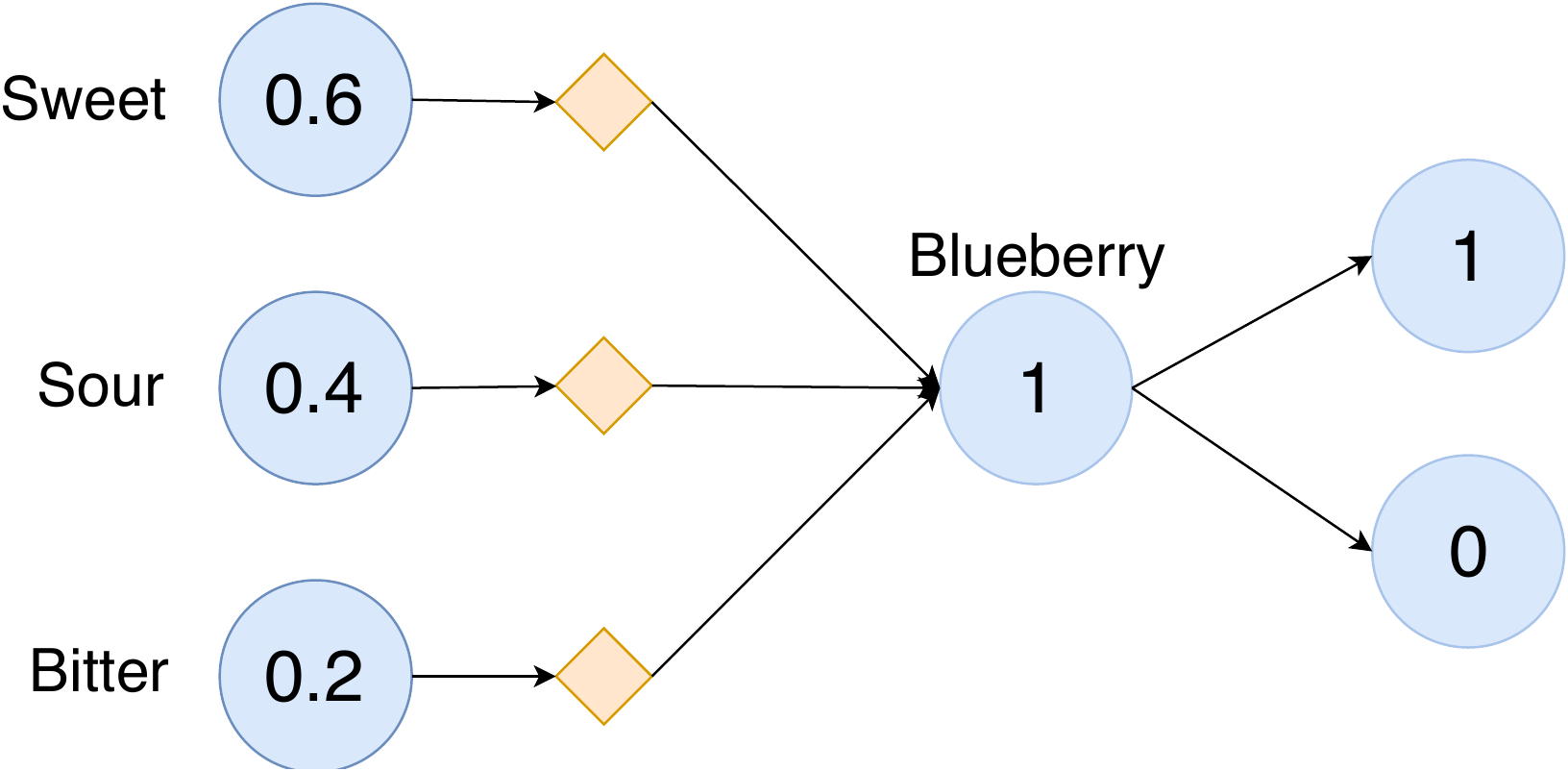}
    \caption{The network shown is created following receipt of the first data point. The node in the center can be viewed as a conjunction node that ``remembers'' the taste of the first berry. Here the numbers represent approximate node activation. 
    }
    \label{fig:berry}
\end{figure}

Imagine an agent learning to distinguish blueberries from blackberries based on taste. Suppose the data points it receives have the form
$(sweetness,\linebreak[1]sourness,\linebreak[1]bitterness;\linebreak[1]blueberry,blackberry)$.
Suppose the first data point is 
$(0.6,0.4,0.2;\linebreak[1]1,0)$. Then LL0 constructs the network shown in Figure \ref{fig:berry}.

\subsection{Generalization}
\begin{figure}[ht]
  \centering
    \includegraphics[width=0.45\textwidth]{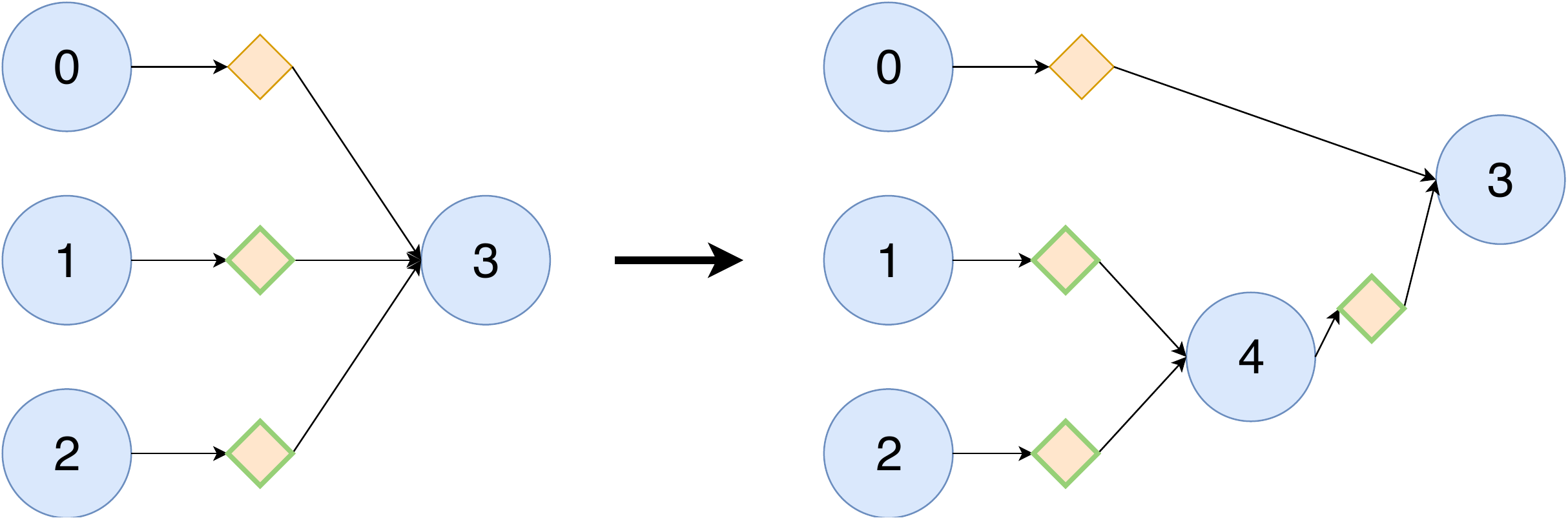}
    \caption{Illustration of the generalization rule. Presuppose the network to the left. Suppose that the value nodes of nodes 1 and 2 are activated, while that of node 0 is not activated. Generalization inserts a new concept node, node 4, as shown to the right.}
    \label{fig:1_gen}
\end{figure}

The generalization rule is used for feature extraction. 
Whenever the extension rule is triggered, LL0 checks whether it can generalize before adding the new node. A concept node $c$ gets generalized if it has enough parent value nodes that are activated above a certain threshold. This is done by detaching the activated parents from $c$ and attaching them to a new intermediate concept node $c'$ that is inserted into the network and connected back to $c$, as illustrated in Figure \ref{fig:1_gen}. The parameters of $c'$ are set so that the original functionality of $c$ is preserved.


\subsection{Forgetting}
The forgetting rule is used for removing relatively unimportant nodes. Forgetting can be done in either of two ways: (i) setting a fixed limit to the network size and removing the worst performing nodes when the network reaches this limit, or (ii) observing how a node performs and removing it if its performance drops below a certain threshold. 
In the case of (i), it can clearly be catastrophic \emph{not} to forget and leave room for new memories. 
The performance $p_c(t)$ of concept node $c$ at time $t$ can be characterized as 
$$ p_c(t) = \frac{\sum_{i=t_0}^t a_i} {t - t_0}, $$
where $t_0$ is the time at which $c$ was added and $a_i$ is the activation of $c$ at time $i$.


\subsection{Backpropagation}
The backpropagation step uses the cross-entropy cost function. 
The partial derivatives are calculated as usual by using the chain rule. Each concept node can be connected to hidden layers further down in the network and to output nodes. The derivative for the concept nodes needs to take all these incoming derivatives into consideration. Three parameters are updated:
\begin{itemize}
    \item bias for the concept node: $\frac{\partial{E}}{\partial{\theta_c}}$;
    \item weights that are not frozen: $\frac{\partial{E}}{\partial{w_i}}$; and
    \item ($\sigma,\mu$) for the value nodes' Gaussian activation function: $\frac{\partial{E}}{\partial{\sigma}}$, $\frac{\partial{E}}{\partial{\mu}}$. 
\end{itemize}
These parameters are multiplied by the learning rate $\delta$ and updated using the gradient-descent algorithm.

\section{Results}
\label{Results}
LL0 was compared to four fully connected, layered networks:
\begin{description}
\item[FC0]: No hidden layer.
\item[FC10]: One hidden layer with 10 nodes.
\item[FC10*2]: Two hidden layers with 10+10 nodes. 
\item[FC10*3]: Three hidden layers with 10+10+10 nodes. 
\end{description}
The hyperparameters of all models were optimized for good overall performance and then fixed. The baseline models were trained using stochastic gradient descent with mini-batch size 10, learning rate 0.01, ReLU nodes in the hidden layers, softmax at the output nodes, and the cross-entropy loss function. 

Despite their simplicity, these static baselines are highly useful. 
Dynamic models that construct fully connected layered architectures generally learn more slowly and consume more energy, since they must search for architectures in addition to undergoing the standard training procedure. 

Performance of LL0 and the four baseline models was analyzed with respect to four data sets adapted from \url{playground.tensorflow.org} and \url{scikit-learn.org}: \textit{spirals}, \textit{digits}, \textit{radiology}, and \textit{wine}, in relation to accuracy and energy consumption on previously unseen test sets. An average over ten runs was calculated for each of the baseline models. Energy consumption for the baseline models was calculated as the number of parameters times the number of forward and backward passes. For LL0 it was calculated similarly and then multiplied by three. 
For the other LL0 rules, energy consumption was estimated conservatively as the number of network parameters times ten.

\subsection{Spirals}
\begin{figure}
\begin{center}
\includegraphics[width=.495\textwidth]{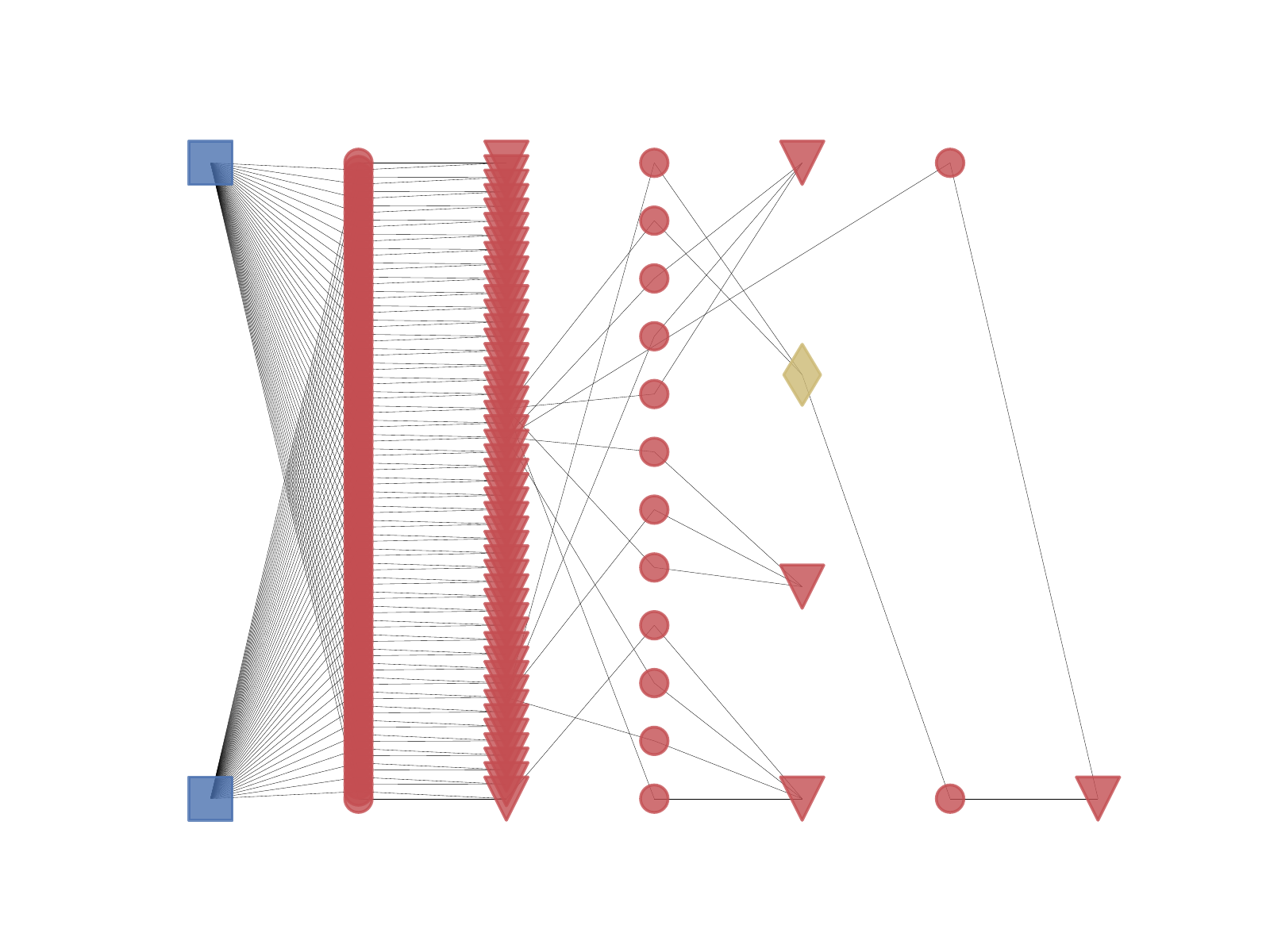}
        \includegraphics[width=.495\textwidth]{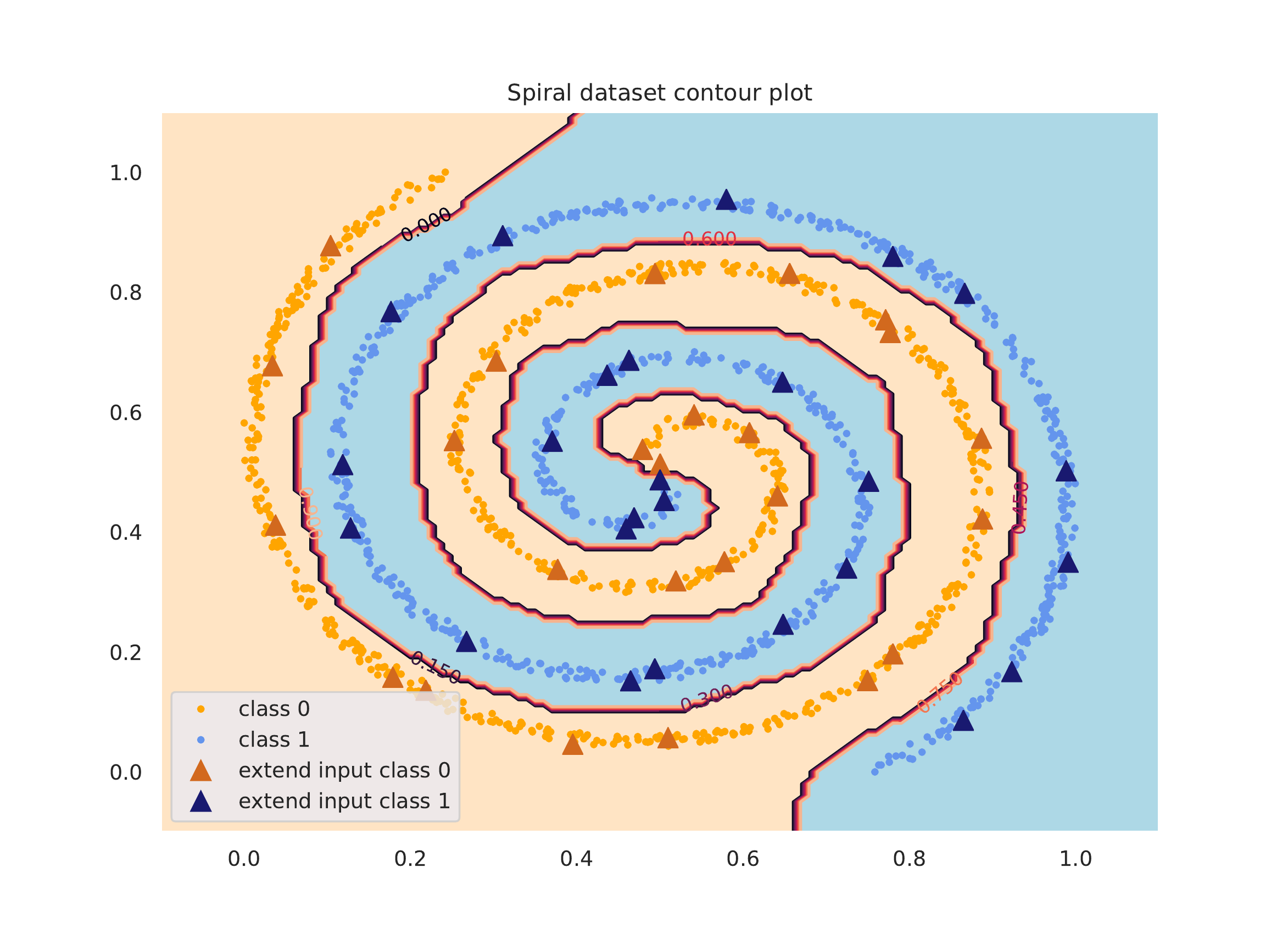}  
        \end{center}
\caption{\textbf{Left}: The network produced by LL0 on the spirals data set, with the two output nodes and their connections omitted for sake of readability. The architecture converged after less than one epoch with about 160 nodes, depth six, and max fan-in five. The yellow node was created by the generalization rule. \textbf{Right}: The spirals data set with the generated decision boundary. Input points that triggered the extension rule are marked by triangles.}

\label{fig:spiralsarch}
\end{figure}

\begin{figure}
\begin{center}
\includegraphics[width=.495\textwidth]{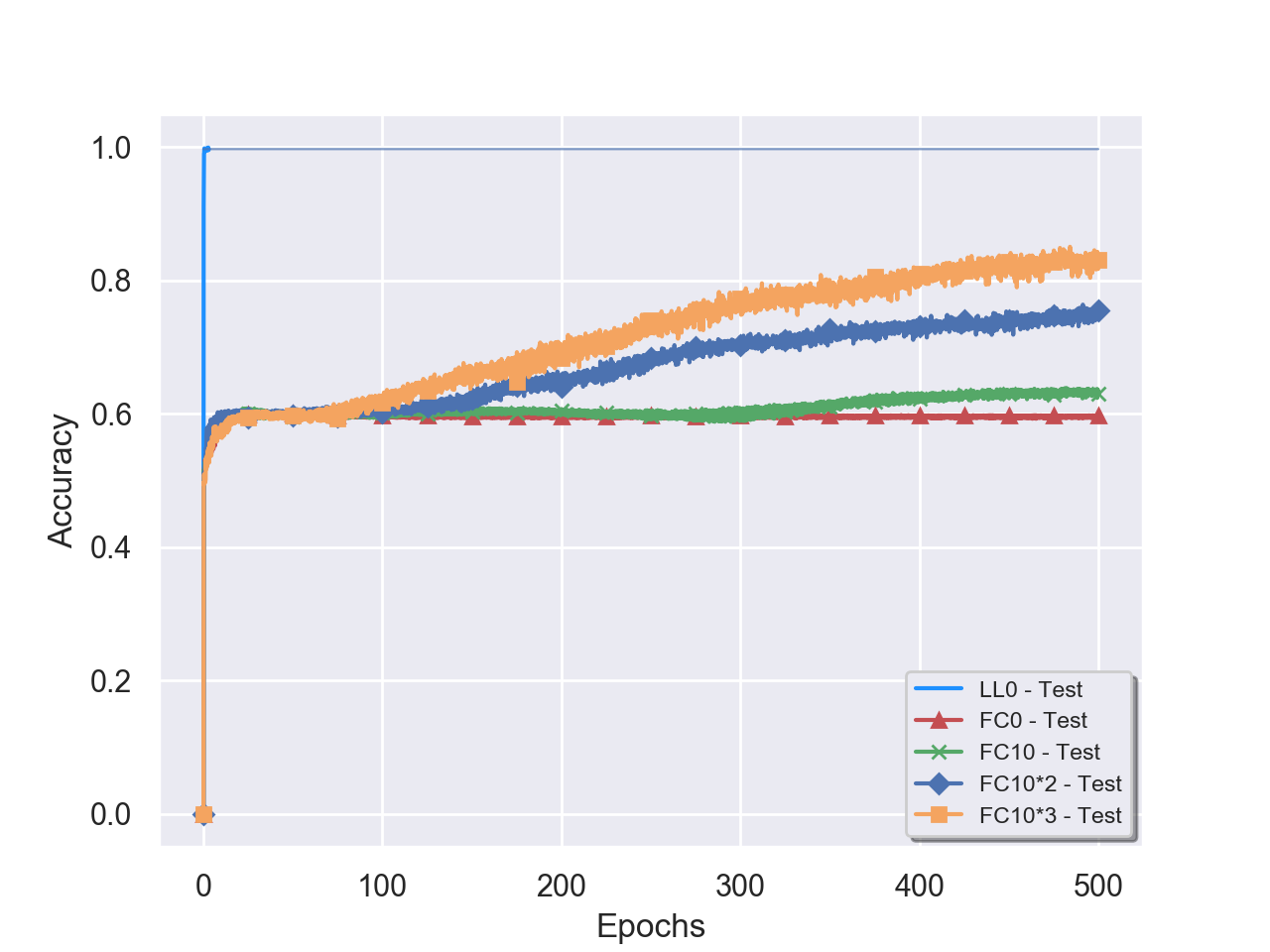}
\includegraphics[width=.495\textwidth]{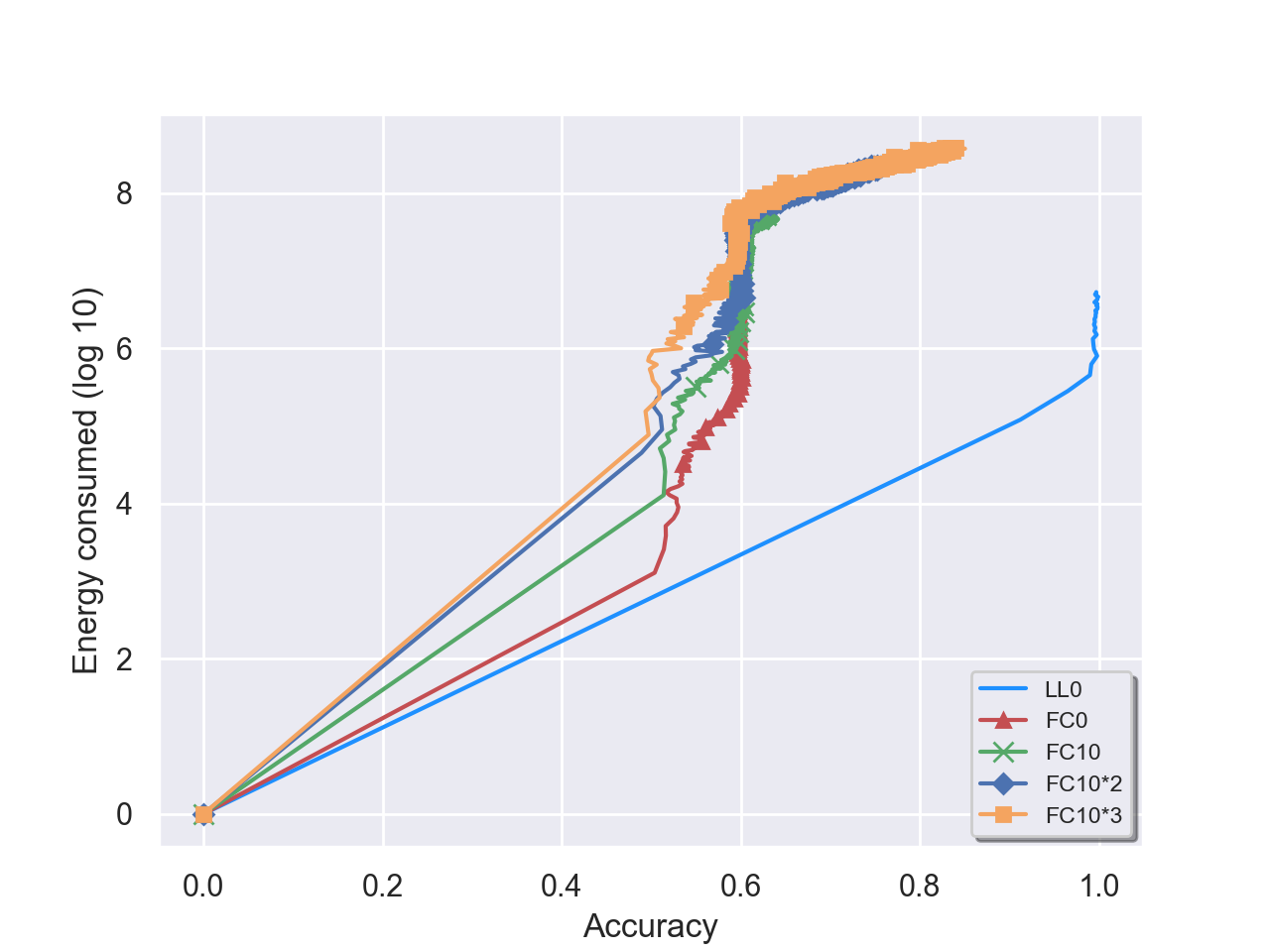}
\end{center}
    \caption{Results on the spirals data set. \textbf{Left}: LL0 reaches 100\% accuracy on the test set after less than one epoch. By contrast, the best baseline model FC10*3 reaches 80\% accuracy after about 350 epochs. \textbf{Right}: FC10*3 consumes over 1000 times more energy than LL0 to reach 80\% accuracy. 
    }
    \label{fig:spirals}
\end{figure}

The \textit{spirals} data set consists of 2,000 two-dimensional data points in the form of two intertwined spirals as shown in Figure \ref{fig:spiralsarch} (right). Figure \ref{fig:spirals} shows the results obtained. 
\noindent

\subsection{Digits}
\begin{figure}
\begin{center}
\includegraphics[width=.495\textwidth]{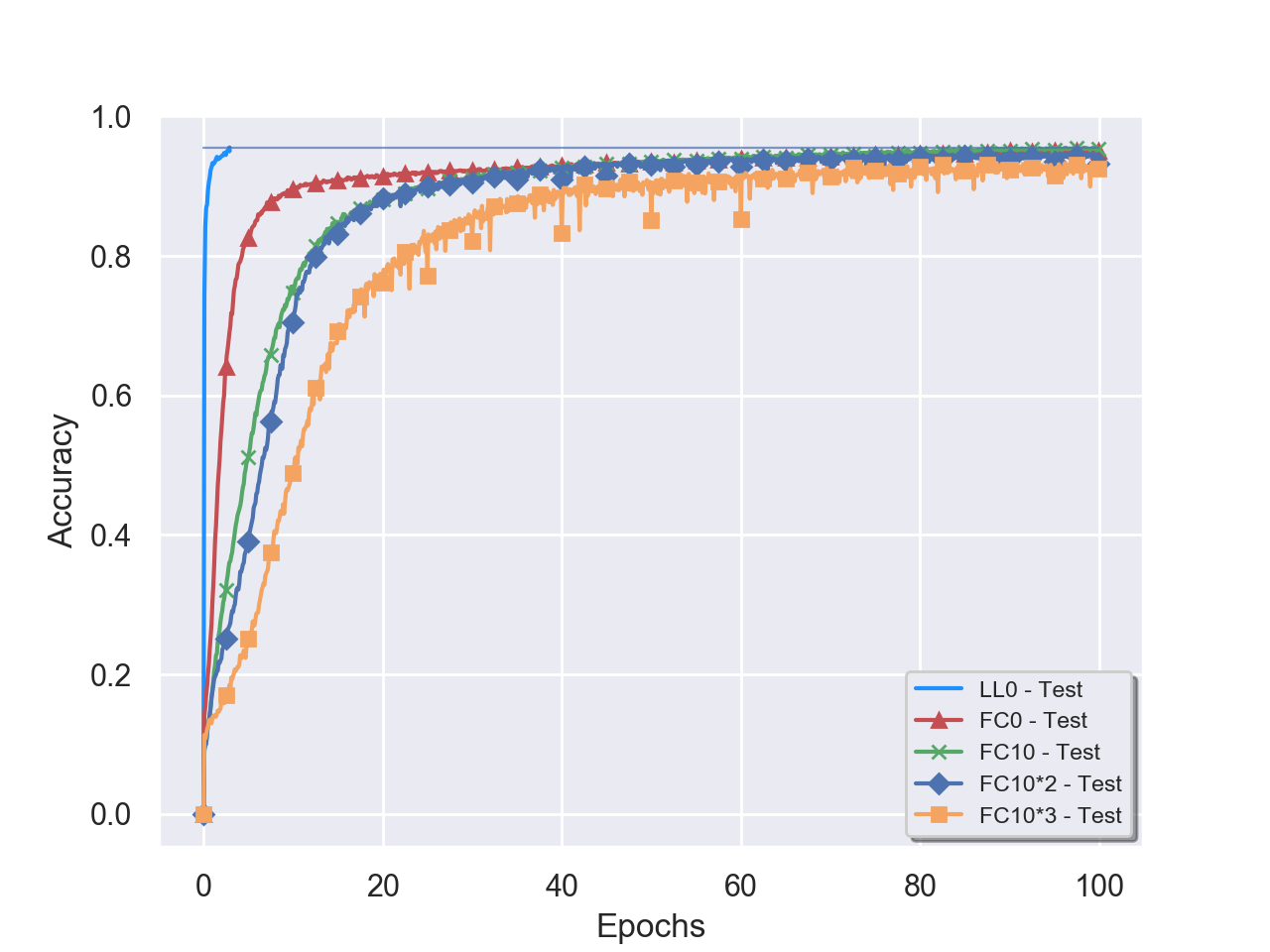}
\includegraphics[width=.495\textwidth]{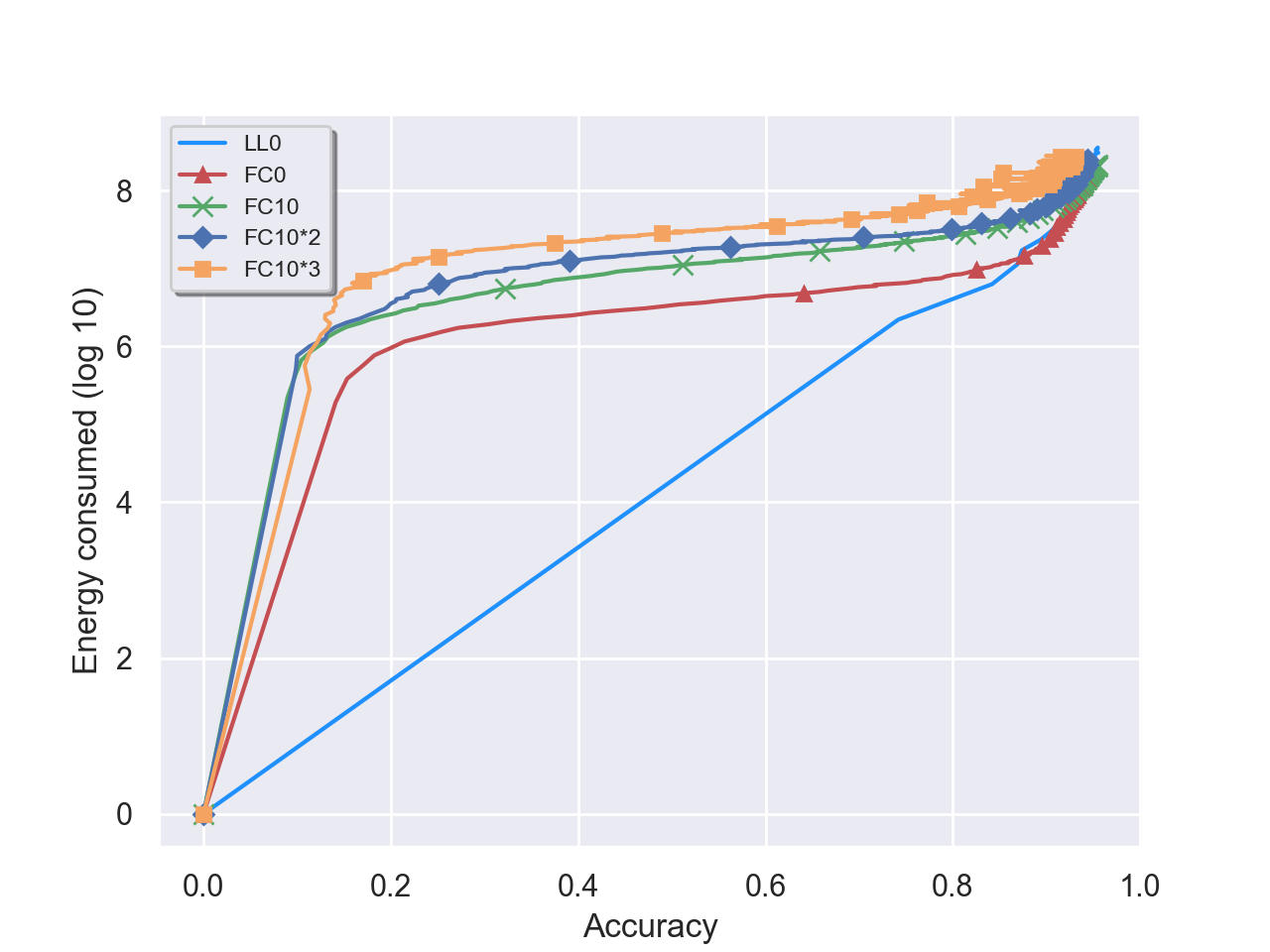}
\end{center}
    \caption{Results on the digits data set. \textbf{Left}: All  models eventually reach approximately the same accuracy. LL0 learns relatively fast. \textbf{Right}: The energy curves converge. 
        }
    \label{fig:digits}
\end{figure}

The \textit{digits} data set consists of 1,797 labeled 8x8 pixel grayscale images of hand-written digits. Figure \ref{fig:digits} shows the results obtained.

\subsection{Radiology}
\begin{figure}
\begin{center}
\includegraphics[width=.495\textwidth]{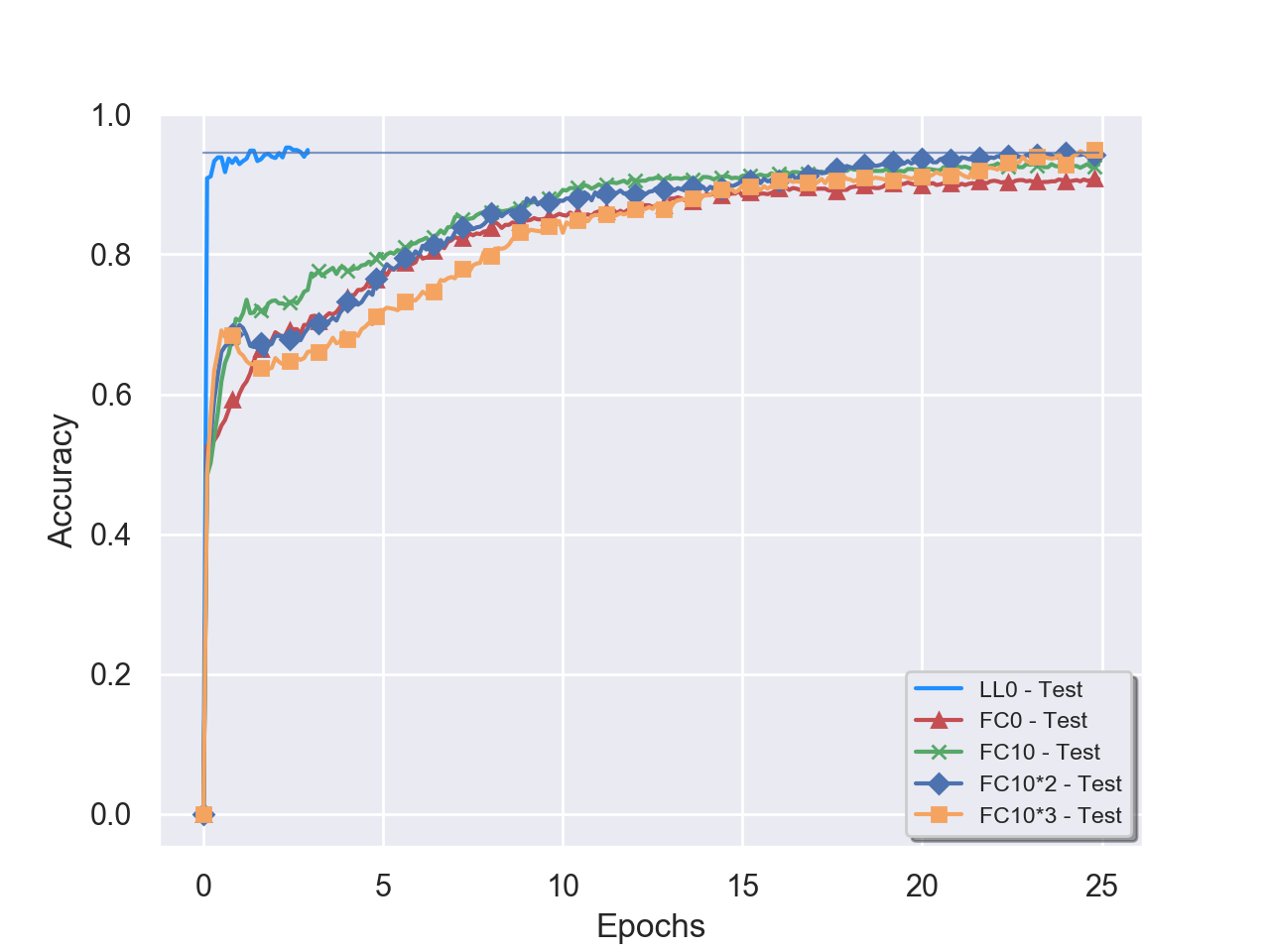}
\includegraphics[width=.495\textwidth]{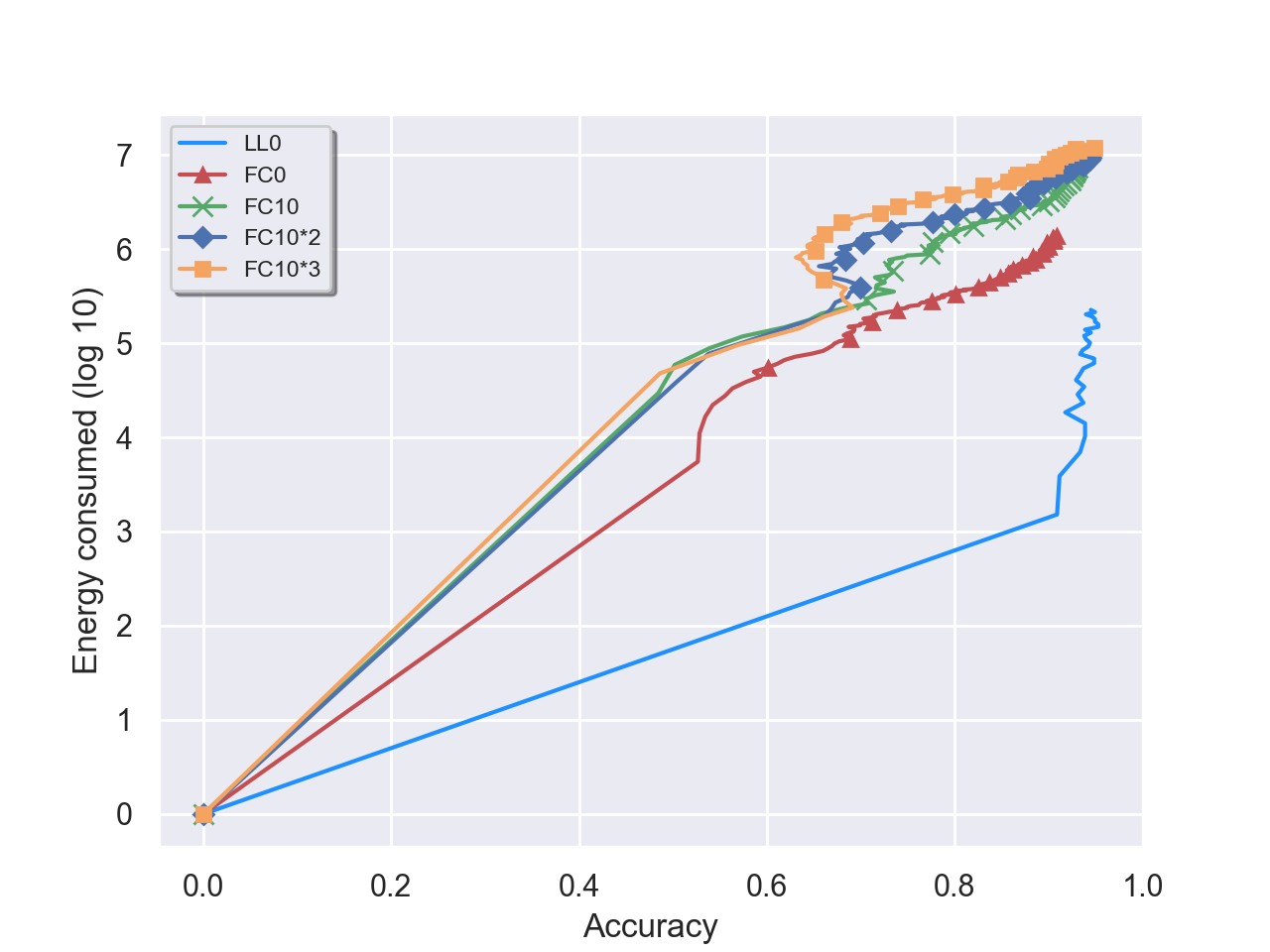}
\end{center}
    \caption{Results on the radiology data set. \textbf{Left}: LL0 learns about ten times faster than the baselines. \textbf{Right}: LL0 consumes about 10\% as much energy. 
    }
    \label{fig:radiology}
\end{figure}

The \textit{radiology} data set consists of 569 data points, each a 30-dimensional vector describing features of a radiology image labeled benign or malignant. Figure \ref{fig:radiology} shows the result obtained.

\subsection{Wine}
\begin{figure}
{%
\centering
\includegraphics[width=.495\textwidth]{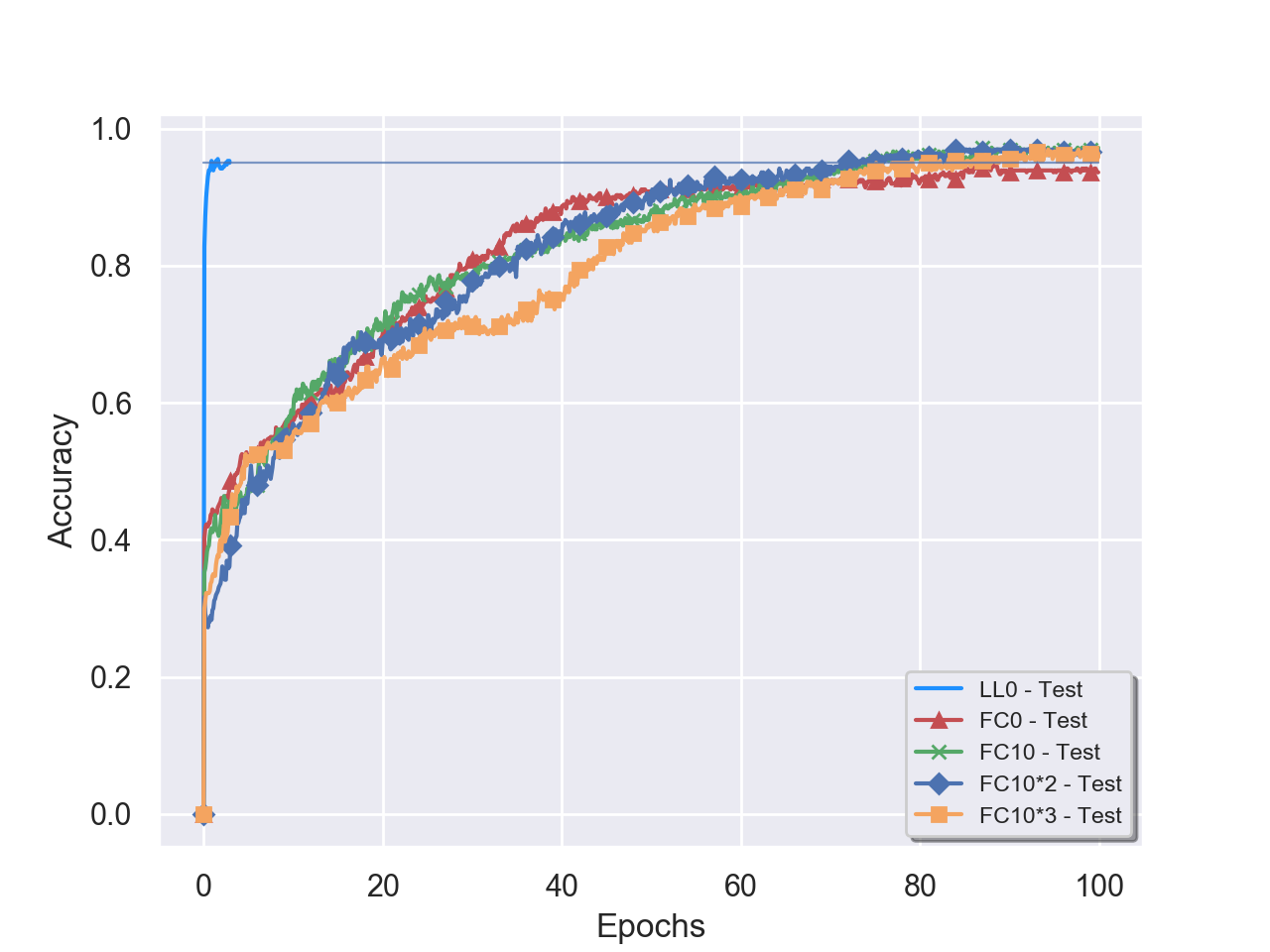}
\includegraphics[width=.495\textwidth]{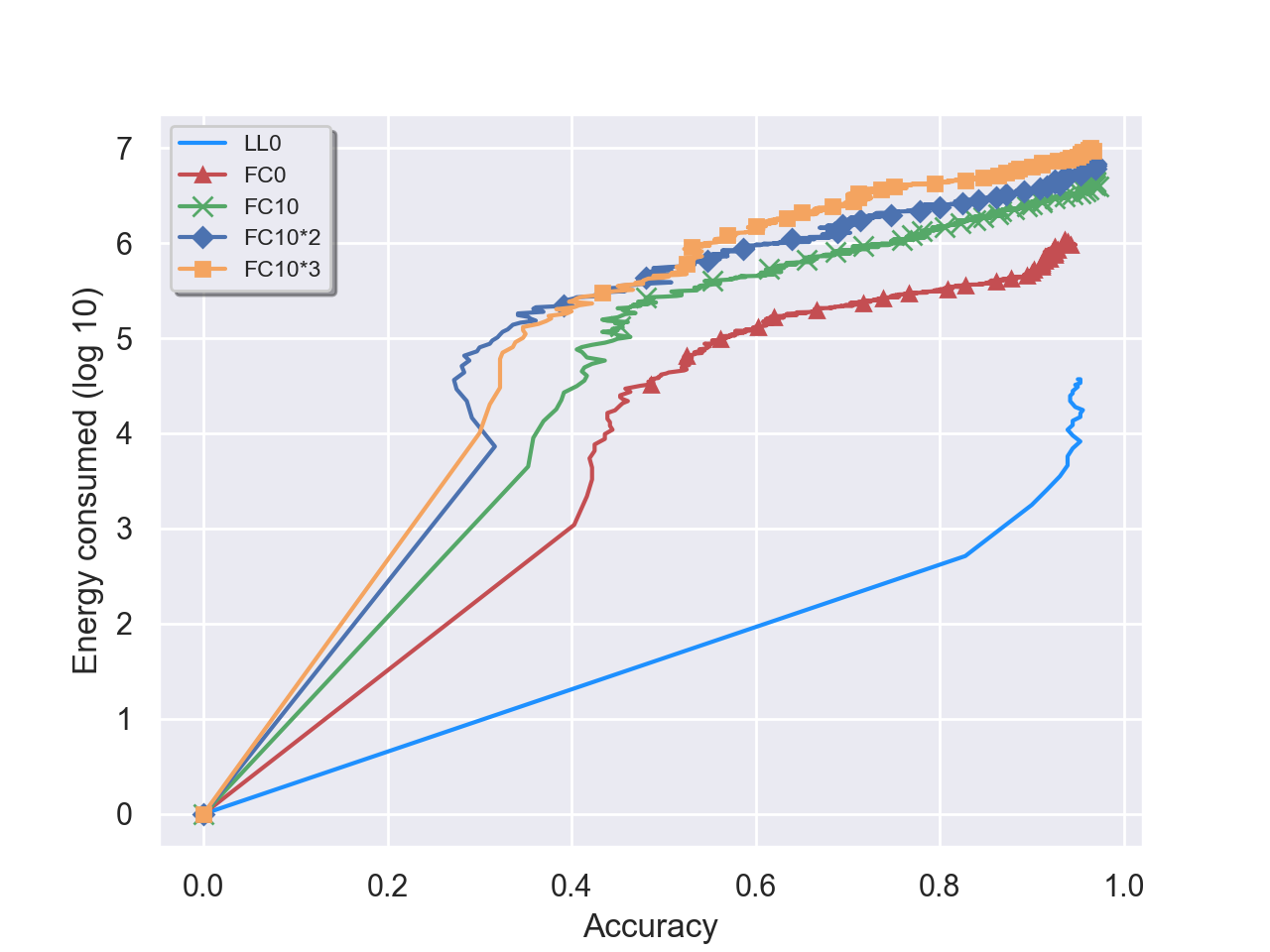}
}
    \caption{Results on the wine data set. \textbf{Left}: LL0 learns much more quickly, but peaks at an accuracy level slightly below the best baseline. \textbf{Right}: Energy consumption.
    }
    \label{fig:wines}
\end{figure}

The \textit{wine} data set consists of 178 data points, each a 13-dimensional vector describing taste features of a wine identified by one of three regions of origin. Figure \ref{fig:wines} shows the results obtained.

\section{Conclusion}
\label{section:Conclusion}
This paper has presented a model for lifelong learning inspired by four types of neuroplasticity. 
The LLO model can be used for constructing networks automatically instead of manually. It starts from a blank slate and develops its deep neural network continuously. It uses no randomization, builds no fully connected layers, and engages in no search among candidate architectures: properties that set it apart from the dynamic models surveyed in Section \ref{Related}. 

The results obtained indicate that LL0 is versatile. The four data sets considered stem from completely different sources: i.e., mathematical functions, handwriting, clinical judgment, and chemical measurements. Still, for each data set, LL0 performs at the level of the best baseline model or better. The reason might be that LL0 uses a form of one-shot learning that counteracts catastrophic forgetting and leads to relatively fast learning and low energy consumption. The fact that LL0 builds sparse networks that are continuously being generalized and trimmed might also play an important role.

The present implementation is a prototype that scales poorly to large data sets although the runtime of the underlying algorithm is linear in the number of nodes. 
Future plans include improving the scalability and extending the model to dynamic deep Q-networks and dynamic recurrent networks. 

\bibliographystyle{splncs04}
\bibliography{main.bbl}

\end{document}